%% file: arxiv_version.tex
\documentclass{article}

 \usepackage[nonatbib,preprint]{nips_2018}

\usepackage[utf8]{inputenc} 
\usepackage[T1]{fontenc}    
\usepackage{hyperref}       
\usepackage{url}            
\usepackage{booktabs}       
\usepackage{amsfonts}       
\usepackage{nicefrac}       
\usepackage{microtype}      
\usepackage{comment}

\usepackage{hyperref}
\usepackage{url}
\usepackage{amsmath}
\usepackage{amssymb}

\def\LSTM{{\textsc{LSTM}}}

\def\LogLoss{{\textsc{LogLoss}}}
\def\softmax{{\text{softmax}}}
\def\maxpool{{\textsc{MaxPool1D}}}

\def\relu{\textsc{Relu}}

\usepackage[T1]{fontenc}

\usepackage{amssymb}
\usepackage{amsmath}
\usepackage{epsfig}
\usepackage{algorithmic}
\usepackage{algorithm}
\usepackage{graphicx}
\usepackage{subfigure}
\usepackage{comment}
\usepackage{url}

\usepackage{xspace}
\usepackage{booktabs} 
\usepackage{amsmath,amssymb,amsfonts}
\usepackage{algorithmic}
\usepackage{graphicx}
\usepackage{textcomp}
\usepackage{amsmath}
\usepackage{algorithm}
\usepackage{color}
\usepackage{comment}

\def\LSTM{{\textsc{LSTM}}}
\def\LogLoss{{\textsc{LogLoss}}}
\def\softmax{{\text{softmax}}}
\def\maxpool{{\textsc{MaxPool1D}}}
\def\relu{{\textsc{ReLU}}}

\def\MPL{LaMP\xspace}
\def\SCL{CPoLS\xspace}

\usepackage{epsfig,endnotes}
\usepackage{times}
\usepackage{url}
\usepackage{epsfig}
\usepackage{graphicx}
\usepackage{subfigure}
\usepackage{amsmath}
\usepackage{amssymb}
\usepackage{amsfonts}
\usepackage{color}
\usepackage{multirow}
\usepackage{algorithm}
\usepackage{algorithmic}
\usepackage[utf8]{inputenc}
\usepackage[T1]{fontenc}
\usepackage{capt-of}
\usepackage{enumerate}
\usepackage{comment}

\usepackage[utf8]{inputenc} 
\usepackage[T1]{fontenc}    
\usepackage{hyperref}       
\usepackage{url}            
\usepackage{booktabs}       
\usepackage{amsfonts}       
\usepackage{nicefrac}       
\usepackage{microtype}      
\usepackage{comment}

\usepackage{hyperref}
\usepackage{url}
\usepackage{amsmath}
\usepackage{amssymb}

\def\LSTM{{\textsc{LSTM}}}

\def\LogLoss{{\textsc{LogLoss}}}
\def\softmax{{\text{softmax}}}
\def\maxpool{{\textsc{MaxPool1D}}}

\def\relu{\textsc{Relu}}
	 {\makeatletter
	\gdef\xxxmark{%
		\expandafter\ifx\csname @mpargs\endcsname\relax 
		\expandafter\ifx\csname @captype\endcsname\relax 
		\marginpar{\textcolor{red}{xxx}}
		\else
		\textcolor{red}{xxx~}
		\fi
		\else
		\textcolor{red}{xxx~}
		\fi}
	\gdef\xxx{\@ifnextchar[\xxx@lab\xxx@nolab}
	\long\gdef\xxx@lab[#1]#2{{\bfseries [\xxxmark \textcolor{red}{#2}
			---{\scshape #1}]}}
	\long\gdef\xxx@nolab#1{{\bfseries [\xxxmark \textcolor{red}{#1}]}}
	\long\gdef\xxx@lab[#1]#2{}\long\gdef\xxx@nolab#1{}%
}

{\makeatletter
	\gdef\addmark{%
		\expandafter\ifx\csname @mpargs\endcsname\relax 
		\expandafter\ifx\csname @captype\endcsname\relax 
		\marginpar{\textcolor{green}{+++}}
		\else
		\textcolor{green}{+++~}
		\fi
		\else
		\textcolor{green}{+++~}
		\fi}
	\gdef\add{\@ifnextchar[\add@lab\add@nolab}
	\long\gdef\add@lab[#1]#2{{\bfseries [\addmark \textcolor{blue}{#2}
			---{\scshape #1}]}}
	\long\gdef\add@nolab#1{{\bfseries [\addmark \textcolor{blue}{#1}]}}
	\long\gdef\add@lab[#1]#2{}\long\gdef\add@nolab#1{}%
}

\title{Robust Neural Malware Detection Models for Emulation Sequence Learning}

\author{Rakshit Agrawal$^{\star}$,
	Jack W. Stokes$^{\dagger}$,
	Mady Marinescu$^{\ddagger}$,
	and~Karthik Selvaraj$^{\ddagger}$\\
	$^{\star}$University of California at Santa Cruz, Santa Cruz, CA 95064 USA\\
	$^{\dagger}$Microsoft Research, One Microsoft Way, Redmond, WA 98052 USA\\
	$^{\ddagger}$Microsoft Corp., One Microsoft Way, Redmond, WA 98052 USA
}

\begin{document}
	
	\maketitle
	
	\input{abstract}

	\input{introduction}

	\input{motivation}
	\input{data}

	\input{threat}
	
	\input{system}

	\input{eval}
	\input{discussion}
	\input{related}
	\input{conc}

	\bibliographystyle{plain}
	\bibliography{kdd2018}
	
\end{document}

%% file: abstract.tex
\begin{abstract}
Malicious software, or malware, presents a continuously evolving challenge in
computer security. These embedded snippets of code in the form of malicious files
or hidden within legitimate files cause a major risk to systems with their ability to run
malicious command sequences.
Malware authors even use polymorphism to reorder these commands and create several
malicious variations. However, if executed in a secure environment, one can perform
early malware detection on emulated command sequences.

The models presented in this paper leverage this sequential
data derived via emulation in order to perform Neural Malware Detection. 
These models target the core of the malicious operation
by learning the presence and pattern of co-occurrence of malicious event actions
from within these sequences.
Our models can capture entire event sequences and
be trained directly using the known target labels. These end-to-end learning models
are powered by two commonly used structures - Long Short-Term Memory (LSTM) Networks
and Convolutional Neural Networks (CNNs).
Previously proposed sequential malware classification models process no more than
200 events. Attackers can evade detection by delaying any malicious activity beyond
the beginning of the file. We present specialized models that can handle extremely
long sequences while successfully performing malware detection in an efficient way.
We present an implementation of the Convoluted Partitioning of Long Sequences approach in order
to tackle this vulnerability and operate on long sequences.
We present our results on a large dataset consisting of 634,249 file sequences,
with extremely long file sequences.
\end{abstract}

%% file: introduction.tex
\section{Introduction}

Malicious software, or malware, is a persistent and continuously growing problem in computer security.
Malware can cause severe issues for computer users. By embedding certain snippets of code within benign
software, it can successfully run malicious commands without being detected by anti-virus software.
This execution can be further modified by malware authors as they can use polymorphism to reorder
malicious commands within different files.
While ensuring coverage of commands relevant to malware, they can hide within large sets of legitimate commands and remain undetected.
Before the file is allowed to be run on the native operating system, the anti-malware engine first
analyzes it using lightweight emulation.
This task of detecting malware within the emulation, however is extremely difficult.
With continuously increasing number of subtle variations observed in malware, simply
constructing sets of rules for detection can become obsolete very quickly. But an underlying fact for any
variant is its dependence on the sequence of commands it needs to operate. While those commands can be
spread out within the execution, they cannot be eliminated or invariably reordered. The nature of their  co-occurrence in sequence or in some patterns is still essential for these variants to operate.

Machine learning models can be trained to learn both the presence and sequential occurrence of events
leading to malicious actions. Neural networks for sequential learning like Long Short-Term
Memory or LSTM~\cite{Hochreiter1997,GersJj1999} recurrent neural networks excel at tasks of learning
sequences with a fixed vocabulary. They have shown exemplary results over the past few years in primary
sequential learning domains like speech~\cite{GravesSpeech2013} and natural language modeling and
translation~\cite{Bahdanau2014,Sutskever}. Besides these,
LSTMs have shown significant success in a much broader range of sequential problems as well. Pointer
Networks~\cite{Vinyals2015a}, Neural Turing Machines and Differentiable Neural
Computers~\cite{Graves2014,Graves2016}, have demonstrated the ability of LSTMs to be used for far
more complex tasks when augmented with attention~\cite{Xu2015} and memory~\cite{Weston2014}.

AI-based learning models for malware detection, therefore, can be constructed with the help of LSTMs
when event sequences are available. Use of language models with Recurrent Neural Networks (RNNs) has been demonstrated by~\cite{PascanuMalware,BenMalware} on malware detection tasks.
The two primary objectives in such problems are event presence detection and sequential occurrence binding.
LSTMs, in particular, are excellent operators for sequential occurrence, and hence lead to significantly
improved results in the process. Presence detection, while done efficiently by RNNs and LSTMs, is a
specialty of the Convolutional Neural Networks, or CNNs~\cite{LeCun1995ConvolutionalNF}. While extremely
effective in computer vision~\cite{krizhevsky2012imagenet,russakovsky2015imagenet}, CNNs
have also shown success on sequential learning problems~\cite{Gehring2016,Gehring2017}. For the closely related
task of multi-class classification among malware families, the authors have shown in~\cite{Kolosnjaji} the use
of CNNs as they capture essential elements across the event sequences.

In this paper, we present robust models for Neural Malware Detection that can operate 
directly on file event sequences in order to learn a probability of the file being malicious in nature. These models are probabilistic, robust and resilient
against polymorphism observed in malicious files.
We present results from these deep models on a large dataset consisting of 634,249 sequences of variable length.

We believe that this is by far the largest study of malware behavioral models.

On performing structural evaluation of these models, we explored a vulnerability persistent even in highly
accurate models that can potentially limit their long term use if the attackers
learn to construct resilient malicious files. All of these models are limited by the length
of the sequences they can operate on, making them ineffective against malware hidden in extremely
long sequences, or malicious files that execute long running loops of legitimate commands before
reaching the malicious events. Potentially, the use of LSTMs allows these models to learn very
long sequences, as the cells can continue to retain context. In
language modeling, this ability helps retain the necessary word context. However in software event
sequences, as observed in malware detection data, context can be reset at several places along the
sequence and a distant hidden event can be harder to correlate with consecutive malicious events. Apart
from that, capturing full length sequences also results in extremely compute-intensive models that
are harder to train and deploy.

In this paper, we present neural models that are immune to the length of the input sequence.
We implement a version of Convoluted Partitioning of Long Sequences (\SCL)~\cite{stokes_scripts} which can capture
sequences of any variable length.
This allows our model to detect malware where events are hidden deep within the sequences.
This model partitions the input sequence and distributes the learning process by using CNNs within
a recurrent learning setting.
We present models for efficiently learning complete event sequences with variable and extremely long
lengths in order to detect malware hidden deep within the sequences. The \SCL 
approach captures the entire length of a sequence by splitting it
into chunks and distributes the learning process by using CNNs in a recurrent fashion.
As shown by Stokes \textit{et al.}~\cite{stokes_scripts}, such models retain the sequential nature of
input entirely without any loss of data.
We also make our models resilient against evasion
by capturing full length sequences without losing any data within the sequence.
\add[Jay]{Might want to change. It allows us to handle
extremely long sequences or loops. in security a vulnerability  is some problem we introduce like a buffer
overflow. In this case, the malware is generating the loop. But I get your point the fixed length
buffer is vulnerable to missing loops, etc. So we might just want to think about it.}

The paper presents a description of all of our models with an emphasis on removing any vulnerability in order to create future resilient Neural Malware Detection systems. We start by discussing the motivation behind this problem and the need for neural models. We then discuss the data.
This is followed by a detailed discussion of our models. Next we describe the experimental process and results obtained on our data. We then present a small discussion on emerging challenges in malware detection, as well as the potential use of our full-length models in other domains. This discussion is followed by a review of related literature. We conclude the paper with a discussion of the benefits of the proposed neural models, and the advantages of \SCL inspired methodologies for performing lossless learning on extremely long sequences.

%% file: motivation.tex
\section{Motivation}
Advances in the space of malware detection using artificial intelligence and
machine learning have significantly strengthened the success rates in detection. Moreover,
the concept of using emulation sequences for file commands provides us with a peek into
internal operations of a malicious file, allowing a learning model to understand its
core. We observe that within these sequences, the primary objective of learning malicious
actions is finding the presence of potential events within entire sequences, and detecting
their sequential occurrence. This seemingly trivial learning task, in reality, is much
more challenging because of the nature of these underlying commands. Individually all
these commands refer to legitimate system operations and cannot be associated with a
level of \emph{badness}. It is the ordered co-occurrence of several such commands which
triggers critical action within system processing to deviate from standard actions.
Numerically, such models can still exhibit a very low error rate, and achieve
a high accuracy. However, due to the nature of such systems and their underlying data, error
rate alone is not a strong representation of a model. Due to their real world impact, false
positive rates in malware detection are also a critical metric of concern since they can
potential prevent a computer from starting or working correctly~\cite{bott_2010}.
A slight miss in the probability measurement
from a malware detection model can turn into a huge cost for the underlying system.

Therefore, it is critical for a good detection model to judge the event sequences
at a much finer granularity and still maintain a general enough weight association with
events to prevent false positives. In order to facilitate such learning, the
training methodology must put into consideration certain characteristics of
the dataset according to the algorithms being used. We believe that neural
models are a good choice because of the sequential nature of the data and
their ability to learn a fixed vocabulary for the set of events observed in
these sequences. An important aspect for a robust Neural Malware Detection
system is to exclude as many assumptions and limitations from the training data
as possible. We suggest using sequences with variation in lengths, with
similar distribution over event vocabulary in both label classes, and
avoiding loss of information throughout the sequences in order to regularize
the dataset sufficiently. We also propose that the model should not be
associated with direct embeddings of these events throughout and must
generate learned tensors in order to escape hard binding of individual
events with malware. It is often possible among training batches to incur
some bias and associate higher weights with certain events commonly
found in sequences leading to a similar label. In such cases, more
significance might be learned towards particular events rather than
learning from entire sequences. A robust model should be able to
maximize detection among sequences, while ensuring generalization
over the entire length of an individual sequence. Such systems can then
learn patterns and ordered co-occurrence of events causing malicious
actions, instead of directly learning event-based activations.

This criticality of the problem and requirement to retain learning within the
respective general blocks motivates our methods for malware detection. Our
models leverage the language model-based approach of LSTMs in order to join the
event sequences, perform convolutions to derive significant occurrences within
extremely long sequences, and ensure end-to-end training in the attempt to learn
embeddings driven directly by the true label. In the next section we discuss our
proposed models in detail and introduce a chunked sequential learning approach
that can help detection in the space of very large sequences hiding critical
information.

%% file: data.tex
\section{Data}
The event sequences that we analyze in this paper are generated by a modified version
of the Microsoft production anti-malware engine which logs the
system API calls made by a portable executable (PE) file (\emph{e.g.}, .exe, .dll).
All of the PE files in our dataset are written for the Microsoft Windows operating system.
To collect
the data, we analyzed a large set of malicious and benign files by scanning
them with the anti-malware engine, and collecting the emulation logs corresponding to each file.
During
the data collection, the malware did not have internet access to prevent the infection of other computers.
The logs were collected in August 2017, and thus represent recent malware families. The length of the
original sequences varies from file to file and is determined by heuristics employed by the anti-malware
engine based on the earlier behavior of the file.

Adversaries sometimes use polymorphism which uses different API calls to achieve the same goal. For
example to  create a file, they may call ofstream.open in C++, fopen in C, or the Windows user mode CreateFile function.
From kernel mode in Windows, they may call the ZwCreateFile or NtCreateFile to create a file.
To handle polymorphism, multiple API calls can be mapped
to the same high-level event, and the total number of high-level events in our data is 156.
The first five events in a sample file are shown in Table~\ref{tab:data}.
In addition, each file is assigned a label $l \in \{0,1\}$ where 1 indicates that file is malware and
a benign file has a label of 0.
\def\arraystretch{1.15}
\begin{table}[]
	\centering
	\caption{Example of the first behavior events in a file}
    \label{tab:data}
	\begin{tabular}{l}
		\hline
		File Event \\ \hline
		createfile \\
        virtualalloc \\
        virtualalloc \\
        getmodulehandle \\
        getmodulefilename \\
	\end{tabular}
\end{table}

The files were randomly selected from our incoming production streams of files to be analyzed and
included in this dataset were selected from a larger collection according to the following
criteria. First, we only included a single file instance for each distinct event sequence.
Second we discarded any file sequences which had multiple labels: we did not include any files with
the same event sequence, but were labeled as both \emph{malware} and \emph{benign}.
Third, we discarded any files with less than 50 events.
The original dataset includes the results of emulating 634,249 files which is then randomly split
into separate training, validation, and test datasets including 443,974, 63,425, and 126,850, respectively.
Distribution over labels in this datasets corresponds to 75\% malicious samples, and 25\% benign samples.

In order to use these sequences with our model, we first create a vocabulary of the 156 event IDs and
then translate the event sequences into our vocabulary space, while keeping a \textit{UNK} entry
for any event outside the vocabulary. 
Each learning batch therefore consists of sequences over
the 156 vocabulary space, resulting in a three dimensional tensor for the batch input. The input dimensions
specify the batch size, sequence length, and event embedding dimension, respectively.

%% file: threat.tex
\section{Threat Model}
The anti-malware engine and its algorithms are under constant attack by adversaries,
and it is important to list the assumptions we make about the attacker with respect
to the proposed system. We first assume that the computer has not been previously infected.
Thus, the trusted computing boundary includes the operating system, user account, and
anti-malware detection system.

Our models analyze data collected from an emulator running in
the anti-malware engine. In some cases, malware tries to detect if it is being executed
in an emulator, and if so, it halts any malicious activity (\textit{i.e.}, cloaking).
Thus, we assume that the malware does not use cloaking to avoid detection.

Since our data is collected in a lab without internet access, we assume that
the malware does not alter its behavior due a failure to communicate
with its command and control infrastructure or other web services.
Before being put into production, the system would need to be trained
and tested with sequences that were obtained with internet access.

%% file: system.tex
\section{System Design}
As discussed previously, our Neural Malware Detection models learn
using the emulation event sequences and are trained end-to-end fashion where
the loss gradient flows directly from the final label. Our objective with these
sequences is to detect the presence of potential events  and their sequential
correlation with other events causing the malicious action. These events can be
both grouped or scattered throughout the sequence. They are not necessarily
consecutive, but are ordered in most cases. Therefore, both presence and order
play an important role in this detection. In this section, we discuss our learning
models in detail. We also comment on certain limitations of our fixed-length models
and present a model that can operate efficiently over sequences of any
length.

\subsection{Building Blocks}
We begin with a short review of the basic units forming our models. In particular, we use
LSTM recurrent neural networks to capture the sequential data. In our extension models,
we also use CNNs to extract significant event occurrences within the entire length of
long sequences. Throughout the models, we also use the concept of Max Pooling along one
dimension. This property helps reduce the dimensionality within the model at different
steps and helps extract significant embeddings and activation wherever required.

\subsubsection{LSTM:}Long Short-Term Memory~\cite{Hochreiter1997,GersJj1999} neural
networks are a memory-based variant of Recurrent Neural Networks (RNNs) where each
neuron is defined to be a gated cell with memory. These networks are less vulnerable
to the vanishing gradient problem~\cite{Hochreiter1998,BengioLongTerm1994} and operate
by maintaining both a hidden state and memory at each time step. This capability of
LSTMs has made them popular in language models and sequential architectures. Among
the different variants of the algorithm that are commonly used, we use the following
equations for our implementation of the LSTM in this paper.

\begin{equation}
	\begin{split}
		& \mathbf i_t = \sigma(\mathbf W_{hi} * \mathbf h_{t-1} + \mathbf  W_{xi} * x_t + \mathbf b_i)\\
		& \mathbf f_t = \sigma(\mathbf  W_{hf} * \mathbf h_{t-1} + \mathbf W_{xf} * x_t + \mathbf b_f) \\
		& \mathbf o_t = \sigma(\mathbf W_{ho} * \mathbf h_{t-1} + \mathbf W_{xo} * x_t + \mathbf b_o) \\
		& \mathbf c_t = \mathbf f_t \odot \mathbf c_{t-1} + \mathbf i_t \odot \tanh(\mathbf  W_{hc} * \mathbf h_{t-1} + \mathbf  W_{xc} * x_t + \mathbf b_c) \\
		& \mathbf h_t = \mathbf o_t \odot \tanh(\mathbf c_t)
	\end{split}
\end{equation}

\noindent where $\sigma$ is the logistic sigmoid function, $\mathbf i_t, \mathbf f_t, \mathbf o_t,
\mathbf c_t$ are the input gate, forget gate, output gate and cell activation respectively. $\mathbf W_h(\cdot)$
are the recurrent weight matrices for each gate, $\mathbf W_x(\cdot)$ are the input weight
matrices per gate, and $\mathbf b(\cdot)$ are the biases for each gate. At each timestep $t$,
the network takes as input a vector $x_t$ and updates both the cell memory $\mathbf c_t$ and
provides a hidden state $\mathbf h_t$. Input vector $x_t$ can be the representation of the input in
any format such as one-hot encoding, or a dense embedding. Function $\odot$ represents the
pairwise product between two vectors. \add[Jay]{xt should be the embedding}

LSTMs are often used in deep models by stacking multiple layers on top of each other. Stacked-LSTMs~\cite{Graves2013b}
allow deeper learning of the structures where each layer gets an embedding from a lower layer while
traversing across the timesteps of the sequence.

\subsubsection{CNN:} Convolutional Neural Networks~\cite{LeCun1995ConvolutionalNF} are extremely powerful models often used in the space of computer vision~\cite{krizhevsky2012imagenet,russakovsky2015imagenet}. Compared to LSTMs, CNNs are faster and more efficient architectures which use smaller kernels that are trained at different locations within the input data. In images, this refers to training smaller blocks within an entire image with the same set of weights instead of using a larger weight tensor for the complete image size. Similarly for one-dimensional data like sentences, CNNs traverse over smaller chunks of the input and perform convolutions across each chunk while updating the smaller weight kernel. Recently, CNNs have shown success on sequential learning problems~\cite{Gehring2016,Gehring2017} and continue to be explored in this new space.

\subsubsection{Max Pooling:} Pooling operations are often used in CNNs~\cite{scherer2010evaluation,maxpool_conv}
to reduce dimensionality and extract significant features in deeper models. Max pooling can
be used with one-dimensional sequences as well allowing for the extraction of higher activations within a deep model. Hence, max pooling is capable of deriving better embeddings for a higher objective than simply obtaining the activation of the final timestep.

\subsection{End-to-End Models}
In order to learn the emulation sequences directly using a target label of malicious behavior, we present our end-to-end models below. For each of these, given an input event sequence $E$ of length $T$ timesteps consisting of events $e_t$ at each time-step $t$ in the sequence and a known label $\tau$, we need to predict the probability $p_m$ of sequence $E$ being malware. In terms of the data under investigation, these models are required to perform two crucial tasks. The first task is the detection of potential events that can lead to a malicious action, and the second is the sequential linkage of events which combine to represent a malicious action. Since these individual events are standard system commands, the mere occurrence of an event within the sequence cannot be used to predict malware. Our initial model designs are inspired by the language model and classifier-based malware detection presented in~\cite{PascanuMalware,BenMalware}

\subsubsection{Direct Sequence Learning}
The Direct Sequence Learning model (\textsc{DSL}) is our most basic end-to-end
architecture. This model relies on the capability of LSTMs to learn sequences
and capture relevant information within the last activation. This model uses
LSTMs as the sequence learning mechanism, which helps convert an entire sequence
into an embedding by using only the last activation from the LSTM. This activation
is then passed through a regular dense layer for learning the derived representation.
We produce the final activation using a logistic sigmoid layer on top which provides
the probability of input sequence being malicious.
Formally, the \textsc{DSL} model is defined as:
\begin{equation}
	\begin{split}
		& h_L = \LSTM(E)[T] \\
		& h_{CL} = \relu(W_L * h_L) \\
		& \mathbf{p_m} = \sigma(W_D * h_{CL})
	\end{split}
\end{equation}
\noindent where $h_L$ is the hidden state activation from the $\LSTM$ at the last
timestep $T$, and $h_{CL}$ is the activation derived from a fully connected $\relu$
neural network layer. $\mathbf{p_m}$ is the final
probability of sequence $E$ being malicious derived through a final logistic sigmoid
layer $\sigma$. $W_L$ is the weight matrix for the $\relu$ layer and $W_D$ is the
weight matrix for the final sigmoid layer for generating the output probability.

\subsubsection{LSTM and Max Pooling}
While \textsc{DSL} is able to translate the sequence into a single vector embedding, it is optimized to predict the last activation. Because of the nature of the  vocabulary in our data, the structure of a sequence is similar to a language model where the objective of the RNN is to predict the next word in the sequence. Therefore, the last activation from the LSTM is best trained for predicting the representation of the next event, and might not be of strong assistance in our objective of finding malicious event sequences.
In~\cite{BenMalware,stokes_scripts}, the authors have used an LSTM and Max Pooling for capturing events from the sequence.
Using a similar LSTM and Max Pooling (\textsc{\MPL}) concept in our models, we perform a temporal max pooling over the resulting LSTM sequence. While in~\cite{BenMalware}, the language model is independently trained using a recurrent neural network, we tie the training of the LSTM with the complete model similar to~\cite{stokes_scripts}. The \textsc{\MPL} model, therefore, first learns the representation for the input sequence events, then performs a temporal max pooling operation on the sequential hidden states to create a final sequence embedding, and then proceeds with dense layers to learn the output probability.
Formally, the \textsc{\MPL} implementation is defined as:
\begin{equation}
	\begin{split}
		& H_L = \LSTM(E) \\
		& h_L = \maxpool(H_L) \\
		& h_{CL} = \relu(W_L * h_L) \\
		& \mathbf{p_m} = \sigma(W_D * h_{CL})
	\end{split}
\end{equation}
\noindent where $H_L$ is the complete sequential output returned by the $\LSTM$, and $\maxpool$ is the temporal max pooling layer that extracts the final embedding $h_L$ from $H_L$.

\subsubsection{Auxiliary-Output Language Learning}
The Auxiliary-Output Language Learning model (\textsc{AoLL}) is a regularized extension of \textsc{\MPL}, built to assist gradient flow in the complete model.
Both \textsc{\MPL} and \textsc{DSL} train the LSTM layer directly from the loss gradient using the final target label. We believe, that while this objective helps direct a specific gradient flow throughout lower layers of the model, it can be assisted by an additional loss in order to address the sequential nature of the data. Models presented in~\cite{PascanuMalware,BenMalware} train a language model over the input sequences where the RNN, at each timestep, is trained to predict the next event in the sequence. In order to incorporate such a goal within the end-to-end learning model, we  use two objectives in our \textsc{AoLL} model. We learn the probability $p_m$ in the same way as presented in \textsc{\MPL}. In addition to this, the \textsc{AoLL} model also obtains an auxiliary output from the LSTM layer in the form of its final activation. We use this output to predict the next event within the sequence, leveraging the sequential learning nature of LSTMs. The complete model is now trained with two objectives, and two loss functions, each of which generates a gradient flow within the model that is used to update the weights.
Formally, \textsc{AoLL} is defined as:
\begin{equation}
	\begin{split}
		& H_L = \LSTM(E) \\
		& h_L = \maxpool(H_L) \\
		& h_{CL} = \relu(W_L * h_L) \\
		& \mathbf{\kappa_L}[t]= \softmax(W_\kappa * H_L[t]) \hspace{16pt} \forall t \in [0,T-1]\\
		& \mathbf{p_m} = \sigma(W_D * h_{CL})
	\end{split}
\end{equation}
\noindent where $\mathbf{\kappa_L}$ is a $\softmax$ output for each timestep of the sequence providing the probability over the entire vocabulary $V$. At each timestep $t$,  $\mathbf{\kappa_L}[t]$ is defined as a vector of size $V$ representing an output probability distribution for each possible event. This measure is aligned with the objective of the language model to predict the next word (event) in the input sequence.
%

As mentioned above, the \textsc{AoLL} model uses two objective functions and therefore requires two loss functions.
For the three models (\textsc{DSL}, \textsc{\MPL}, and \textsc{AoLL}) defined above, we use Log Loss as our loss function. For determining the probability $p_m$, we use a loss in the form of binary cross-entropy. For the stepwise outputs $\mathbf{\kappa_L}$ in \textsc{AoLL}, we use categorical cross-entropy as the loss function when predicting the next event in the sequence.
Formally, for all three models, with prediction $p_m$ and known target label $\tau$, we measure the final loss $\mathcal{L}$ as
\begin{equation}
\mathcal{L} = \LogLoss(\mathbf{p_m}, \tau).
\end{equation}
For \textsc{AoLL}, with predictions $\mathbf{\kappa_L}$ and events $e_t$ in the event sequence $E$ for timestep $t$, we measure an auxiliary loss $\mathcal{L}_{aux}$ as
\begin{equation}
	\begin{split}
		&\mathcal{L}_{aux} = \LogLoss(\mathbf{\kappa_L}[t], e_{t+1}) \hspace{16pt}  \forall t \in [0,T-1].
	\end{split}
\end{equation}

\subsection{Model Limitations}
The models defined above are all trained end-to-end using a known target label and input event sequences only. A significant strength of these models is their ability to learn the sequence embeddings that are critical informants of the malicious behavior in the sequence caused by the consecutive or sequential occurrence of certain events. While the use of the LSTM provides us with the ability to use variable length sequences, it often becomes computationally very expensive to train as the lengths of the sequences increase. In general sequence learning, it is often common to cut the sequences up to a certain prescribed length and then use those fixed length subsequences for training. In non end-to-end models, it is possible to train the sequences on a certain sequence length and then use a different length for further representation when used with deeper models. However these solutions are still limited by the computation and memory capacities.

In language model based applications, capturing a limited length of the input sequence can often yield a sufficient representation of the complete sequence. However in the case of detection within a longer sequence, our objective is to find all the events that can lead to a malicious action, and they can be separated to reach very long distances within the entire sequence. For instance, consider a malicious event that requires events $v_a$, $v_b$, and $v_c$ $\in V$ to happen sequentially but does not need them to occur consecutively. This event sequence, therefore, can incorporate a large number of random events $v_{r1}$ \ldots $v_{rn}$ between $v_a$ and $v_b$, and $v_b$ and $v_c$ respectively. An optimal system needs to detect the presence of these events and generate an appropriate activation on detecting their sequential occurrence. If the number of random events $n$ increases drastically, the model can lose context of presence of $v_a$ by the time it reaches $v_b$. One approach to retain distant event occurrences is by using Bidirectional LSTMs~\cite{Graves2013b}, but in very long sequences this becomes even more computationally expensive for the end-to-end model to train.

Another problem with limited length sequences in this case is that malware can often
be written as a long series of legitimate benign events followed by malicious events
much later in the sequence (\emph{i.e.}, file). Limiting the length of such sequences
provides potentially benign sequences to the system with a malicious label. Therefore,
for training on limited length, we need to filter data for prevention against such cases,
leading to loss of data. When given a very large dataset with malicious events occurring
within the smaller length sequences, a model can be trained well. However, limiting the
length of sequences adds a vulnerability to the model for the detection of
malware. It is therefore essential for a model to capture the entire
length of the sequences in order to find the events located at larger distances and
still maintain their sequential order. \add[Jay]{Do our models capture the entire length of the sequence? Or does it again require a fixed length of fixed length sequences?}

\subsection{Convoluted Partitioning of Long Sequences}
In order to capture both the presence of events, and their sequential relationship
in very long sequences, we use the Convoluted Partitioning of Long Sequences (\SCL) approach
introduced in~\cite{stokes_scripts} for detecting malicious JavaScript and VBScript.
A similar approach is presented in~\cite{nvidia_malware_exe} where the authors
performed a static analysis on the chunked PE file byte sequence. These models were not effective 
in either case.
The implementation adopted in~\cite{nvidia_malware_exe} uses the last hidden output from the
LSTM. However based on our experimental results with RNNs and LSTMs,
temporal max pooling often performs significantly better than using the last hidden state when performing event detection within a sequence.
We believe that this model approach applies well to our problem of malware detection since the length of
event sequences can be extremely long.
While the authors in~\cite{stokes_scripts} have demonstrated the use of \SCL on shorter byte sequences of length 1000,
we extend this model to any length and do not use any limiting factors in our models.
In our \SCL models, we are able to learn end-to-end models from the entire length of very long
sequences and capture the sequential nature of events.
We formalize the \SCL model in Algorithm~\ref{alg:sequentialchunkedlearning}.

For an input event sequence $E$, we first split it into a chunked sequence $C=[c_1, c_2, c_3, \ldots]$, where $E=c_1|c_2|c_3|\ldots$, using Algorithm~\ref{alg:chunkify}. Since $E$ itself is a sequence, subsequences $c_i \in C$ are ordered in nature. This operation increases the dimensionality of the batch input tensor from 3 to 4 dimensions. We now treat each chunk as an element of our top-level sequence which is passed through a recurrent or a time distributed layer that sequentially processes each chunk. Within each of these recurrent processes, we perform convolutional learning on the subsequence within that timestep as shown in Algorithm~\ref{alg:recurrentconvolutions}. By performing convolutions, our goal is to extract significant event presence from each subsequence and reduce the overall dimensionality of the problem. Therefore, each chunk $c_i$, in the recurrent processing, is passed through a 1-dimensional convolutional neural network (\textsc{Conv1D}), and then through a time distributed max pooling layer (\maxpool). This allows us to reduce the size of each chunk, and of the entire sequence. We recombine the outputs of each chunk to form a new sequence representing activations from recurrent convolutions.
We then pass this derived sequence through the LSTM of our end-to-end models. Due to the usage of the derived sequences, we cannot use the \textsc{AoLL} model with \SCL.

\begin{algorithm}[tb]
	\caption{\textsc{Convoluted Partitioning of Long Sequences}}
	\label{alg:sequentialchunkedlearning}
	\begin{algorithmic}
		\STATE {\bfseries Input:} Sequence $E$, Chunk Size $s$
		\STATE $C$ = \textsc{Chunkify}($E$, $s$)
		\STATE $H_{RC}$ = \textsc{RecurrentConvolutions}($C$)
		\STATE $E'$ = $[h_{RC1}, h_{RC2}, h_{RC3} \ldots]$
		\STATE $\mathbf{p_m}$ = \textsc{\MPL}($E'$)
		\RETURN  $\mathbf{p_m}$
	\end{algorithmic}
\end{algorithm}

\begin{algorithm}
		\caption{\textsc{Chunkify}}
		\label{alg:chunkify}
		\begin{algorithmic}
			\STATE {\bfseries Input:} Sequence $E$, Chunk Size $s$
			\STATE Initialize $chunks = [\hspace{2pt}]$.
			\FOR{$i=0$ {\bfseries to} $len(E)/s$}
			\STATE \textsc{Append}($E[s*i:(s+1)*i]$, $chunks$)
			\ENDFOR
			\RETURN $chunks$
		\end{algorithmic}

\end{algorithm}

\begin{algorithm}[t]
	
	\caption{\textsc{RecurrentConvolutions}}
	\label{alg:recurrentconvolutions}
	\begin{algorithmic}
		\STATE {\bfseries Input:} Chunks $C$
		\FOR{$c_k$ in $C$}
		\STATE $h_{Ck}$ = \textsc{Conv1D}($c_k$)
		\STATE $h_{MPk}$ = \maxpool($h_{Ck}$)
		\ENDFOR
		\STATE $H_{MP}$ = $[h_{MP1}, h_{MP2}, h_{MP3}, \ldots]$
		\RETURN $H_{MP}$
	\end{algorithmic}
\end{algorithm}

%% file: eval.tex
\section{Experiments and Results}
We performed an extensive evaluation of the models presented above with our emulation sequence dataset.
We implemented these models on the Tensorflow~\cite{TensorflowShort} platform using the Keras~\cite{keras} library. Both
models and data access operations were written using the Python programming language.
Our models were trained on 2 Nvidia Tesla K40m GPUs.

For each model, we ran several iterations to identify the best hyper-parameter settings.
We used LSTM as our recurrent neural network module in each model. The hidden dimension of 1500 was used for the LSTM
in each model. For our vocabulary of size 156, we used an embedding dimension of size 114 throughout the models.
The $\relu$ layer used in our models had a hidden dimension of 64. The CNN layer used in the \SCL model performed one-dimensional convolutions on a window of 10 items, with a stride size of 5.
The CNNs used a channel size of 114, which is equal to the embedding dimension, in order to consider each dimension within the computation.
Each model was run for 15 epochs on 443,974 training samples, and validated on 63,425 samples after each epoch.
For the limited length models, we used sequences of length 256.
We used the \textsc{Adam} optimizer~\cite{Kingma2014} with a learning rate of 0.001. For the limited length models we used
a mini-batch size of 64. For full length models, we used a mini-batch size of 32.
The results presented in this section use 126,850 test samples on our trained models.
For each model, we present the average results over multiple iterations using the best settings.

\begin{figure}[ht]
	\vskip 0.00005in
	\begin{center}
		\centerline{\includegraphics[width=0.9\columnwidth]{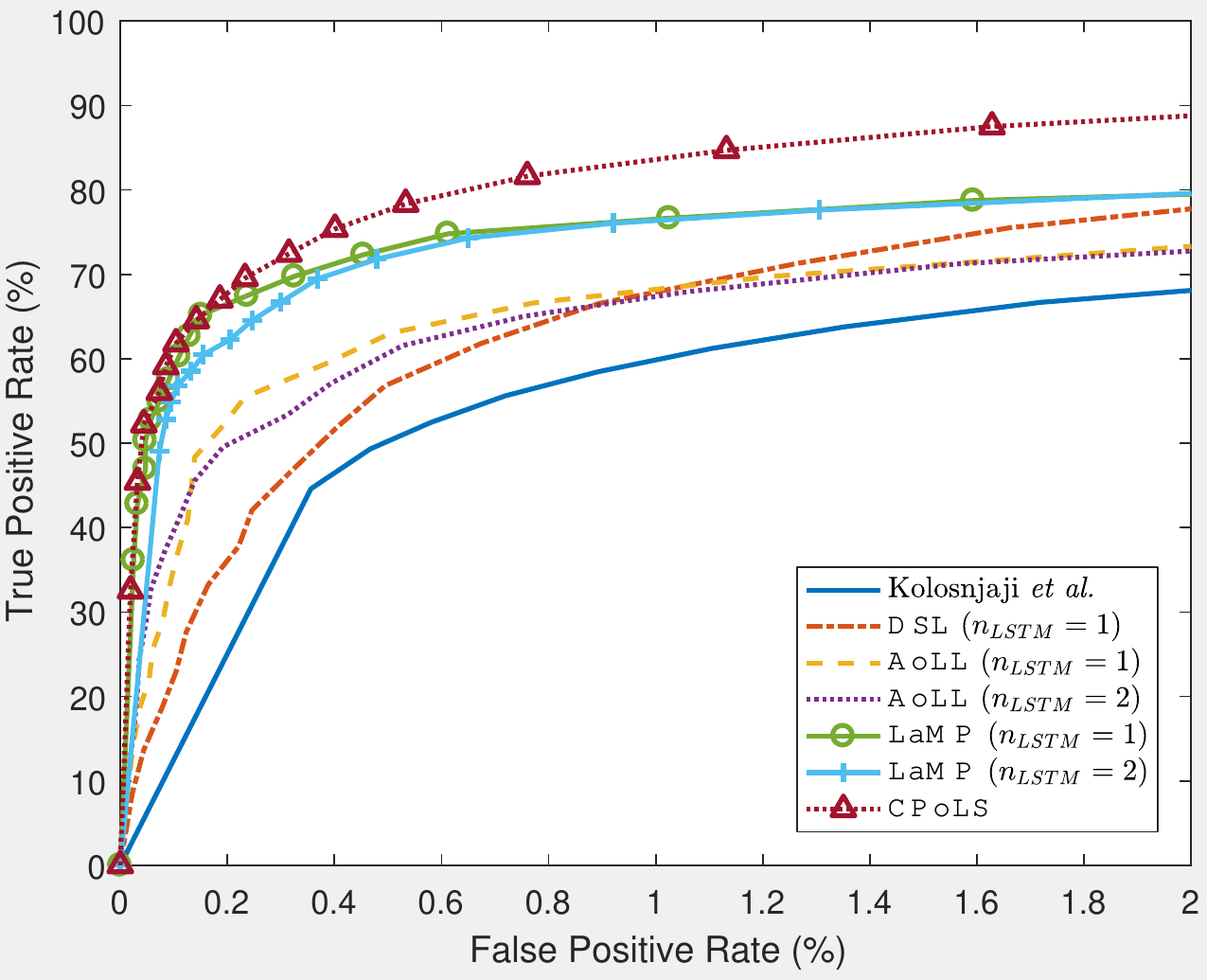}}
		\caption{Receiver operating characteristic plots for the models described above}
		\label{fig:ROC}
	\end{center}
	\vskip -0.00005in
\end{figure}

We present results across three significant metrics. We first discuss the average result accuracies
derived by our models in predicting the probability $p_m$ of a file being malicious. Accuracies
provide us with an overall strength of the Neural Malware Detection systems in detecting malicious
action within a sequence.
As we discussed earlier, this problem is more critical and sensitive to false positives due to the nature of the underlying systems.
Therefore, we also compare our models using Receiver Operating Characteristic (ROC)
plots, in order to get a finer sense of each model's performance.
Along with the ROC plot, we also measure the True Positive Rate (TPR) at fine scale in order to learn the performance of
our system. We measure the TPR, therefore at a False Positive Rate (FPR) of 1\%.
Table~\ref{tab:accuracies} summarizes the accuracies and TPRs across models and configurations.
Figure ~\ref{fig:ROC} presents the ROC plots for the best performing version of each model.
We have limited the x-axis scale on the ROC plots to a finer scale of 2\% on the FPR.
By visualizing at an FPR of 2\%, we evaluate each model's performance at a
much finer scale than possible through accuracies.
Even with minor variation in accuracies, it can be observed that the \SCL model operating on full length sequences
performs much better at a very low FPR as well, maximizing the effective area under the curve.

As can be seen, \SCL not only builds resilience against long sequences,
but also performs significantly better than the other models
on our dataset. These neural models respond aptly to the criticality of our problem. The model
which combines a CNN followed by an LSTM proposed by Kolosnjaji, et al.~\cite{Kolosnjaji}
performed better than \textsc{DSL} in terms of average accuracy, but had a significantly lower performance
in TPR@1\% and the best case ROC curve.
Our experiments with \textsc{AoLL} display comparable performance
to \textsc{\MPL}~\cite{PascanuMalware,BenMalware}, highlighting the ability of end-to-end models to identify
optimal gradient flow even when using multiple loss functions, but also signifying the effectiveness of a single loss function
optimizing the entire model end-to-end.
These results also confirm the ability to train models
with supervised (classification) and unsupervised (language learning) methods simultaneously in an end-to-end setting.

In our experiments with full length models, we also tested different chunk sizes and observed
that the selection of chunk size did not
significantly affect the results, but did increase training speeds. We experimented with
different sizes between 32 and 256, but could not obtain significant accuracy variation.
This might be due to the fact that each of these chunk sizes were sufficient to capture
important event occurrence and can be subject to further evaluation in the future. The speed of
training, however, improved with larger chunks going under convolutions, hence lowering the
sequence lengths for the LSTMs at the following stages.

\def\arraystretch{1.15}
\begin{table}[]
	\centering
	\caption{Test accuracies (\%) and TPRs (\%) at an FPR = 1\% for all the models}
\label{tab:accuracies}
	\begin{tabular}{lll}
		\textsc{Model Name} &\textsc{ Accuracy}	&\textsc{ TPR@1\% }\\ \hline \hline
		DSL  $n_{LSTM}$=1  & 0.881	&  68.44    \\ \hline
		Kolosnjaji et al.	& 0.932	& 59.96		\\ \hline
		AoLL  $n_{LSTM}$=1 & 0.922	& 68.14    \\
		AoLL  $n_{LSTM}$=2 & 0.932	& 67.40    \\ \hline
		\MPL  $n_{LSTM}$=1  & 0.947	& 76.50     \\
		\MPL  $n_{LSTM}$=2  & 0.951	& 76.27    \\ \hline
		\SCL  $n_{LSTM}$=1  & 0.956	& 83.50
	\end{tabular}
\end{table}

%% file: discussion.tex
\section{Discussion}
It is important to consider evasion when considering machine learning models
for security applications. Papernot, et al.~\cite{Papernot2016} recently proposed an adversarial attack
for recurrent neural networks. Recurrent models are more challenging to attack
than deep neural networks due their recurrent nature. To attack a recurrent structure,
the authors first unfold (\textit{i.e.}, unroll) the recurrent model into essentially a deep
neural network with many layers. Next they compute the unfolded Jacobian and perturbations
to craft adversarial sequences which can be used to trick the RNN into predicting the
incorrect class.

To the best of our knowledge, no adversarial defenses have been proposed for
recurrent neural networks. Thus, defenses against adversarial attacks directed
at recurrent models, such as those proposed in this paper, are an open research topic.

Besides malware detection, the models presented in this paper also provide potential
approaches for handling extremely large sequences in general. Neural network based models
are applied to a large number of applications across several domains. Depending on the nature
of the respective data, the length of sequences in these domains can play a significant role in the model's
learning capability. While we presented our models for malware detection, our architecture, in particular
is not tied to the underlying problem and can be generalized for a broader set of problems. Learning
on sequences involving event detection, event prediction or any related objective can benefit from
our models in efficiently using the entire length of the sequences.
We believe, with certain domain-specific modifications, and with more generalization within the overall
architecture, our models can help perform lossless sequential learning irrespective of data lengths. 

%% file: related.tex
\section{Related Work}
The persistent increase in the growth and distribution of malware has called
for the alarming need of detection and prevention. Anti-virus and anti-malware
systems developed over time have performed in-depth analysis of malicious
software families, have developed signatures, tagging methods and several
other mechanisms to tackle this ever growing problem. Exploration of using
machine learning in this space has witnessed the use of both traditional and
deep learning models. Support Vector Machines (SVM) have been incorporated
into this task by~\cite{Pfoh2013}. Hidden Markov Models have been explored
by~\cite{Attaluri2009}. Assistance in signature generation using deep learning
has been demonstrated in~\cite{DavidDBN} where they used Deep Belief Networks
to generate malware signatures that could then be passed through classification
models. Entering more into the space of neural network-based models, language structure within emulated
sequences of a malicious executable have been utilized by~\cite{PascanuMalware,BenMalware}. They have built
RNN-based language models to learn the relationship between the event structure and malware.

The \SCL model used in this paper is based on~\cite{stokes_scripts} and uses CNNs over sequential
structures. While popularly used in image and graphical data, CNNs have also been used
by~\cite{Gehring2016},~\cite{Gehring2017}, where they have shown exceptional results
in the space of language models by constructing Sequence-to-Sequence models using
CNNs. In malware classification,~\cite{Kolosnjaji2017} and \cite{Kolosnjaji} also demonstrated the use of CNNs
along with sequential learning. In their models for malware
classification, they demonstrated the ability to classify among malicious families
from a dataset of malicious files. Among other applications, an interesting mix of
RNN and CNNs has been presented by~\cite{7552276} by first using RNNs for feature extraction and
generating image representations from files which are classified further by CNNs.

%% file: conc.tex
\section{Conclusions}
Neural Malware Detection models, as presented in this paper, help combat
the challenge of detecting malware from extremely long API call sequences
generated through an Anti-malware engine. While helping to resolve a major
concern in computer security, our models also address the critical and sensitive nature
of this problem.

Through this paper, we presented several end-to-end models that use combinations
of LSTM recurrent neural networks and CNNs. Our models learn entirely from event sequences
while learning event embeddings within the deep models themselves.
Through our full-length based models, we present efficient methods to perform lossless
sequence learning on extremely long sequences in detection tasks. Through our training and
results on the largest malware dataset of 634,249 sequences, we demonstrated the significance
of using entire sequence lengths when performing malware detection.

In our \textsc{AoLL} model, we have also demonstrated the use
of multiple objectives and losses within an end-to-end setting that utilize additional gradient flows within
the model while maximizing the primary objective of predicting the probability of maliciousness.
Through \SCL, we demonstrated the advantages
of using Convolutional Neural Networks for event detection in association with LSTMs for sequential binding.
Through our \textsc{Chunkify} approach, we presented an effective method for the break down of extremely long sequences in learning tasks without losing the sequential nature at any stage of the model.

In summary, this paper targets the core of the malware operation style and learns from it a
Neural Malware Detection model, which can be used directly with emulators, can be embedded
within anti-malware systems to serve as a detection operator, can efficiently handle long sequences, and is resilient to variations and loops emerging in malware over time.

%% file: arxiv_version.bbl
\begin{thebibliography}{10}

\bibitem{TensorflowShort}
Mart\'{\i}n Abadi, Ashish Agarwal, Paul Barham, et~al.
\newblock {TensorFlow}: Large-scale machine learning on heterogeneous systems,
  2015.
\newblock Software available from tensorflow.org.

\bibitem{BenMalware}
B.~Athiwaratkun and J.~W. Stokes.
\newblock Malware classification with lstm and gru language models and a
  character-level cnn.
\newblock In {\em 2017 IEEE International Conference on Acoustics, Speech and
  Signal Processing (ICASSP)}, pages 2482--2486, March 2017.

\bibitem{Attaluri2009}
Srilatha Attaluri, Scott McGhee, and Mark Stamp.
\newblock Profile hidden markov models and metamorphic virus detection.
\newblock {\em Journal in Computer Virology}, 5(2):151--169, May 2009.

\bibitem{Bahdanau2014}
Dzmitry Bahdanau, Kyunghyun Cho, and Yoshua Bengio.
\newblock {Neural Machine Translation by Jointly Learning to Align and
  Translate}.
\newblock sep 2014.

\bibitem{BengioLongTerm1994}
Y.~Bengio, P.~Simard, and P.~Frasconi.
\newblock Learning long-term dependencies with gradient descent is difficult.
\newblock {\em IEEE Transactions on Neural Networks}, 5(2):157--166, Mar 1994.

\bibitem{bott_2010}
Ed~Bott.
\newblock Defective mcafee update causes worldwide meltdown of xp pcs, Apr
  2010.

\bibitem{keras}
Fran\c{c}ois Chollet et~al.
\newblock Keras.
\newblock \url{https://github.com/fchollet/keras}, 2015.

\bibitem{maxpool_conv}
Dan~C. Cire\c{s}an, Ueli Meier, Jonathan Masci, Luca~M. Gambardella, and
  J\"{u}rgen Schmidhuber.
\newblock Flexible, high performance convolutional neural networks for image
  classification.
\newblock In {\em Proceedings of the Twenty-Second International Joint
  Conference on Artificial Intelligence - Volume Volume Two}, IJCAI'11, pages
  1237--1242. AAAI Press, 2011.

\bibitem{DavidDBN}
Omid~E David and Nathan~S Netanyahu.
\newblock {DeepSign: Deep Learning for Automatic Malware Signature Generation
  and Classification *}.

\bibitem{Gehring2016}
Jonas Gehring, Michael Auli, David Grangier, and Yann~N. Dauphin.
\newblock {A Convolutional Encoder Model for Neural Machine Translation}.
\newblock nov 2016.

\bibitem{Gehring2017}
Jonas Gehring, Michael Auli, David Grangier, Denis Yarats, and Yann~N. Dauphin.
\newblock {Convolutional Sequence to Sequence Learning}.
\newblock may 2017.

\bibitem{GersJj1999}
Felix~A {Gers Jj}, Urgen Schmidhuber, and Fred Cummins.
\newblock {Learning to Forget: Continual Prediction with LSTM}.
\newblock 1999.

\bibitem{Graves2013b}
Alex Graves, Navdeep Jaitly, and Abdel-rahman Mohamed.
\newblock {Hybrid speech recognition with Deep Bidirectional LSTM}.
\newblock In {\em 2013 IEEE Workshop on Automatic Speech Recognition and
  Understanding}, pages 273--278. IEEE, dec 2013.

\bibitem{GravesSpeech2013}
Alex Graves, Abdel{-}rahman Mohamed, and Geoffrey~E. Hinton.
\newblock Speech recognition with deep recurrent neural networks.
\newblock {\em CoRR}, abs/1303.5778, 2013.

\bibitem{Graves2016}
Alex Graves, Greg Wayne, and Et~al.
\newblock {Hybrid computing using a neural network with dynamic external
  memory}.
\newblock {\em Nature (in Press)}, 538(1), 2016.

\bibitem{Graves2014}
Alex Graves, Greg Wayne, and Ivo Danihelka.
\newblock {Neural Turing Machines}.
\newblock oct 2014.

\bibitem{Hochreiter1998}
Sepp Hochreiter.
\newblock The vanishing gradient problem during learning recurrent neural nets
  and problem solutions.
\newblock {\em Int. J. Uncertain. Fuzziness Knowl.-Based Syst.}, 6(2):107--116,
  April 1998.

\bibitem{Hochreiter1997}
Sepp Hochreiter and Jurgen Schmidhuber.
\newblock {Long short-term memory}.
\newblock {\em Neural Computation}, 9(8):1--32, 1997.

\bibitem{Kingma2014}
Diederik~P. Kingma and Jimmy Ba.
\newblock {Adam: A Method for Stochastic Optimization}.
\newblock dec 2014.

\bibitem{Kolosnjaji2017}
Bojan Kolosnjaji, Ghadir Eraisha, George Webster, Apostolis Zarras, and Claudia
  Eckert.
\newblock {Empowering convolutional networks for malware classification and
  analysis}.
\newblock In {\em 2017 International Joint Conference on Neural Networks
  (IJCNN)}, pages 3838--3845. IEEE, may 2017.

\bibitem{Kolosnjaji}
Bojan Kolosnjaji, Apostolis Zarras, George Webster, and Claudia Eckert.
\newblock Deep learning for classification of malware system call sequences.
\newblock In {\em Australasian Joint Conference on Artificial Intelligence},
  pages 137--149. Springer International Publishing, 2016.

\bibitem{krizhevsky2012imagenet}
Alex Krizhevsky, Ilya Sutskever, and Geoffrey~E Hinton.
\newblock Imagenet classification with deep convolutional neural networks.
\newblock In {\em Advances in neural information processing systems}, pages
  1097--1105, 2012.

\bibitem{LeCun1995ConvolutionalNF}
Yann LeCun and Yoshua Bengio.
\newblock Convolutional networks for images speech and time series.
\newblock 1995.

\bibitem{Papernot2016}
Nicolas Papernot, Patrick McDaniel, Ananthram Swami, and Richard Harang.
\newblock Crafting adversarial input sequences for recurrent neural networks.
\newblock 2016.

\bibitem{PascanuMalware}
R.~Pascanu, J.~W. Stokes, H.~Sanossian, M.~Marinescu, and A.~Thomas.
\newblock Malware classification with recurrent networks.
\newblock In {\em 2015 IEEE International Conference on Acoustics, Speech and
  Signal Processing (ICASSP)}, pages 1916--1920, April 2015.

\bibitem{Pfoh2013}
Jonas Pfoh, Christian Schneider, and Claudia Eckert.
\newblock {\em Leveraging String Kernels for Malware Detection}, pages
  206--219.
\newblock Springer Berlin Heidelberg, Berlin, Heidelberg, 2013.

\bibitem{nvidia_malware_exe}
E.~{Raff}, J.~{Barker}, J.~{Sylvester}, R.~{Brandon}, B.~{Catanzaro}, and
  C.~{Nicholas}.
\newblock {Malware Detection by Eating a Whole EXE}.
\newblock {\em ArXiv e-prints}, October 2017.

\bibitem{russakovsky2015imagenet}
Olga Russakovsky, Jia Deng, Hao Su, Jonathan Krause, Sanjeev Satheesh, Sean Ma,
  Zhiheng Huang, Andrej Karpathy, Aditya Khosla, Michael Bernstein, et~al.
\newblock Imagenet large scale visual recognition challenge.
\newblock {\em International Journal of Computer Vision}, 115(3):211--252,
  2015.

\bibitem{scherer2010evaluation}
Dominik Scherer, Andreas M{\"u}ller, and Sven Behnke.
\newblock Evaluation of pooling operations in convolutional architectures for
  object recognition.
\newblock {\em Artificial Neural Networks--ICANN 2010}, pages 92--101, 2010.

\bibitem{stokes_scripts}
Jack~W. Stokes, Rakshit Agrawal, and Geoff McDonald.
\newblock Neural classification of malicious scripts: A study with javascript
  and vbscript.
\newblock {\em CoRR}, 2018.

\bibitem{Sutskever}
Ilya Sutskever, Oriol Vinyals, and Quoc~V Le.
\newblock {Sequence to Sequence Learning with Neural Networks}.

\bibitem{7552276}
S.~Tobiyama, Y.~Yamaguchi, H.~Shimada, T.~Ikuse, and T.~Yagi.
\newblock Malware detection with deep neural network using process behavior.
\newblock In {\em 2016 IEEE 40th Annual Computer Software and Applications
  Conference (COMPSAC)}, volume~2, pages 577--582, June 2016.

\bibitem{Vinyals2015a}
Oriol Vinyals, Meire Fortunato, and Navdeep Jaitly.
\newblock {Pointer Networks}.
\newblock jun 2015.

\bibitem{Weston2014}
Jason Weston, Sumit Chopra, and Antoine Bordes.
\newblock {Memory Networks}.
\newblock oct 2014.

\bibitem{Xu2015}
Kelvin Xu, Jimmy Ba, Ryan Kiros, Kyunghyun Cho, Aaron Courville, Ruslan
  Salakhutdinov, Richard Zemel, and Yoshua Bengio.
\newblock {Show, Attend and Tell: Neural Image Caption Generation with Visual
  Attention}.
\newblock feb 2015.

\end{thebibliography}
